\documentclass[10pt,twocolumn,letterpaper]{article}

\usepackage[pagenumbers]{cvpr} 
\usepackage{algorithm}
\usepackage{algpseudocode}
\usepackage{booktabs}
\usepackage{multirow}
\usepackage{makecell}
\usepackage{adjustbox}
\usepackage[accsupp]{axessibility}

\definecolor{cvprblue}{rgb}{0.21,0.49,0.74}
\usepackage[pagebackref,breaklinks,colorlinks,allcolors=cvprblue]{hyperref}

\def\paperID{FedVision-17} 
\def\confName{CVPR}
\def\confYear{2026}

\author{Sina Gholami\thanks{Equal contribution}\\
University of North Carolina at Charlotte\\
Charlotte, NC, USA\\
{\tt\small sina.gholami72@gmail.com}
\and
Abdulmoneam Ali\footnotemark[1]\\
University of North Carolina at Charlotte\\
Charlotte, NC, USA\\
{\tt\small aali28@charlotte.edu}
\and
Tania Haghighi\\
University of North Carolina at Charlotte\\
Charlotte, NC, USA\\
{\tt\small thaghigh@charlotte.edu}
\and
Ahmed Arafa\\
University of North Carolina at Charlotte\\
Charlotte, NC, USA\\
{\tt\small aarafa@charlotte.edu}
\and
Minhaj Nur Alam\\
University of North Carolina at Charlotte\\
Charlotte, NC, USA\\
{\tt\small minhaj.alam@charlotte.edu}
}

\title{FedSIR: Spectral Client Identification and Relabeling for Federated Learning with Noisy Labels}

\begin{document}
\maketitle
\begin{abstract}

Federated learning (FL) enables collaborative model training without sharing raw data; however, the presence of noisy labels across distributed clients can severely degrade the learning performance. In this paper, we propose \textbf{FedSIR}, a multi-stage framework for robust FL under noisy labels. Different from existing approaches that mainly rely on designing noise-tolerant loss functions or exploiting loss dynamics during training, our method leverages the spectral structure of client feature representations to identify and mitigate label noise. 

Our framework consists of three key components. First, we identify clean and noisy clients by analyzing the spectral consistency of class-wise feature subspaces with minimal communication overhead. Second, clean clients provide spectral references that enable noisy clients to relabel potentially corrupted samples using both dominant class directions and residual subspaces. Third, we employ a noise-aware training strategy that integrates logit-adjusted loss, knowledge distillation, and distance-aware aggregation to further stabilize federated optimization. Extensive experiments on standard FL benchmarks demonstrate that \textbf{FedSIR} consistently outperforms state-of-the-art methods for FL with noisy labels. The code is available at \url{https://github.com/sinagh72/FedSIR}.  

\end{abstract}

\section{Introduction}
\label{sec:intro}


With the rapid growth of data-generation sources, federated learning (FL) has emerged as a promising paradigm for preserving privacy by keeping data at its sources while enabling collaborative model training across distributed clients \cite{mcmahan}. However, the effectiveness of collaborative learning still depends on the quality of locally stored data. In practice, client-side datasets are often collected in uncontrolled environments and may contain noisy or unreliable labels, which can significantly degrade the performance of federated models \cite{RCC-ali}.

To address the challenge of noisy labels, a large body of literature has proposed various strategies. A representative overview of related work is provided in Section~\ref{sec:related}. Most existing methods rely on signals derived from loss trajectories or client performance during training \cite{fedcorr}. However, such signals may become unreliable in FL settings due to inherent FL challenges, including data heterogeneity, aggregation strategies, and partial client participation. 

In \cite{ClipFL}, the authors leverage the availability of a clean validation dataset at the central server to identify clean clients and prune noisy ones. Instead of excluding potentially noisy clients, \cite{fedned} incorporates them through knowledge distillation (KD), in which noisy client models act as negative teachers that guide the global model away from their predictions. In \cite{RHFL}, a different aggregation strategy is proposed to mitigate the impact of noisy clients. In addition to the challenge of identifying noisy clients, most existing label correction strategies rely on the predictions of the trained model.    

As a result, many FL approaches for noisy labels heavily depend on training dynamics, making it difficult to disentangle the impact of FL system characteristics from the effect of noisy labels. This motivates our work to approach the noisy-label problem from a different perspective by asking the following question:  
\begin{center}
\textit{Can structural information in the feature representations of local datasets help distinguish clean samples from corrupted ones?}
\end{center} 
We answer this question in the affirmative. Our proposed framework demonstrates that the spectral structure of class-wise feature representations provides a reliable signal for identifying noisy clients and correcting potentially corrupted labels in FL under completely random label noise, i.e., symmetric label noise.

\noindent\textbf{Contributions.}
The main contributions of this work are summarized as follows:
\begin{itemize}

    \item We propose \textbf{FedSIR}, a multi-stage FL framework that addresses noisy labels through the \emph{spectral geometry of client feature representations}, providing a new perspective beyond conventional loss-dynamics or prediction-based approaches.

    \item We introduce a \textbf{spectral client identification} mechanism that characterizes each client through class-wise feature subspaces and their off-diagonal similarity statistics, enabling the identification of clean and noisy clients with minimal communication overhead and without requiring access to raw data, or a noise transition model.

    \item We develop a \textbf{spectral relabeling} scheme for noisy clients that constructs class references from clean-client dominant directions and residual subspaces, and performs conservative label correction by enforcing agreements between complementary spectral criteria.

    \item We integrate noise-aware federated optimization, combining logit-adjusted (LA) loss, KD, and distance-aware aggregation (DaAgg) to further stabilize training under label noise.  

    \item We conduct extensive experiments on federated CIFAR-10 datasets under varying label-noise rates and client heterogeneity levels, showing that \textbf{FedSIR} consistently outperforms strong baselines and recent methods for FL with noisy labels.
\end{itemize}

\textbf{Notation.} We use bold uppercase letters to denote matrices, bold lowercase letters to denote vectors, and non-bold symbols to denote scalars. For a set $\mathcal{C}$, $|\mathcal{C}|$ denotes its cardinality. For a vector $\mathbf{x}$, $\|\mathbf{x}\|_2$ denotes the $\ell_2$ norm.

\section{Related Works}
\label{sec:related}
\noindent \textbf{Noisy label learning in centralized setting.} In centralized learning, where all training data are aggregated and processed on a central server, multiple methods have been proposed to address the noisy label challenge. One prominent category is sample selection and reweighting, which aims to identify clean samples during training. In \cite{co-teaching}, two neural networks are trained simultaneously, and each network selects small-loss samples to update its peer. To mitigate the risk that both networks converge to the same erroneous predictions, \cite{co-teaching_plus} further introduces a disagreement-update mechanism before the cross-update. 

Another line of work focuses on designing robust loss functions. For example, \cite{wang2019symmetric} proposes the symmetric cross entropy loss, which augments standard cross entropy with a reverse cross entropy term to improve robustness without requiring explicit noise estimation. Similarly, \cite{ma2020normalized} proposes a normalization framework that provably converts any loss function into a noise-robust variant and introduces the active passive loss, which combines normalized active and passive components to balance learnability and robustness. 

In addition, different training strategies have been explored to limit the impact of noisy labels,  including Mixup \cite{zhang2018mixup}, Dividemix \cite{li2020dividemix}, KD methods \cite{bhardwaj2021knowledge, SD, jiang2023knowledge}, and early-learning regularization techniques \cite{liu2020early}.

Despite their effectiveness, these methods assume centralized access to training data. As a result, additional adaptations may be required to apply them in FL settings where data is distributed across multiple clients, often exhibiting strong heterogeneity. Furthermore, clients may contain different levels of label noise, and the model is trained through collaborative aggregation rather than centralized optimization. These characteristics introduce new challenges for noisy label learning in federated environments, and motivate the study of noisy label learning in FL settings.

\begin{figure*}[t]
  \centering
  \includegraphics[width=\textwidth]{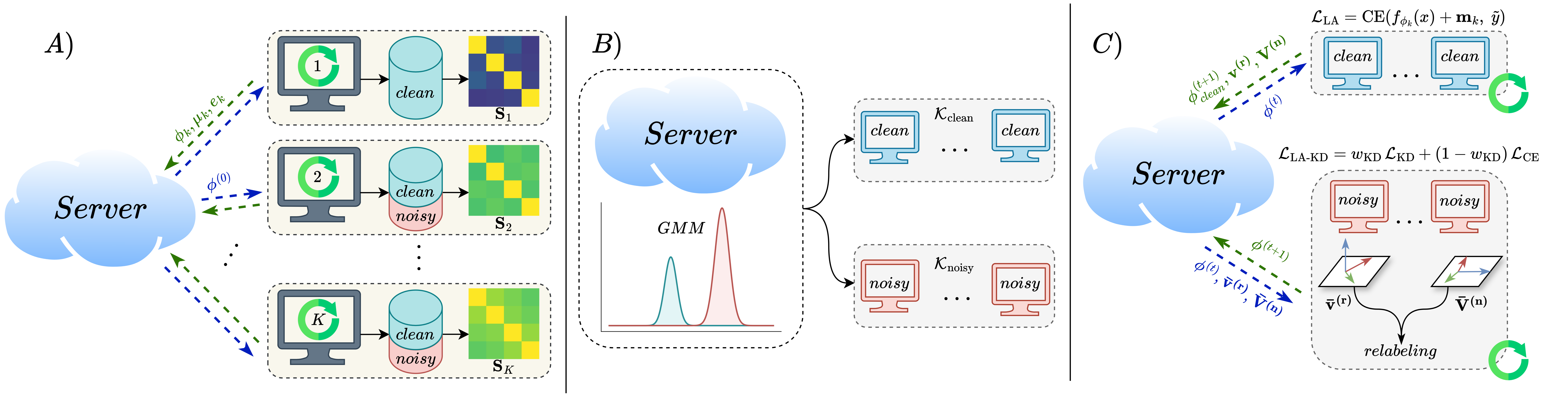}
  \caption{\textbf{A)} Each client trains the global model over its local data (noisy and clean), computes its class similarity matrix and its off-diagonal mean and energy. The client then sends the off-diagonal mean and energy, along with updated gradients, to the server. \textbf{B)}
  The server fits GMM to the off-diagonal statistics derived from class similarity matrices of all clients and partitions the clients into two groups: noisy and clean. \textbf{C)} The server aggregates the class-wise dominant direction $\mathbf{\bar{v}}^{(r)}$ and residual subspaces $\mathbf{\bar{V}}^{(n)}$ from clean clients, and sends them to noisy clients with a clean reference model for relabeling.
  }
  \label{fig:pipeline}
\end{figure*}

\noindent \textbf{Noisy label learning in FL.} RoFL \cite{rofl} adapts the sample selection paradigm to the federated setting by leveraging globally aggregated class-wise feature centroids. Each client uses these centroids to identify confident samples for training and applies pseudo-labeling to the remaining noisy instances. Moving beyond sample-level selection, RHFL \cite{RHFL} introduces a client confidence re-weighting scheme that adaptively adjusts each client’s aggregation weight based on its estimated label quality and learning efficiency. In contrast, FedCorr \cite{fedcorr} identifies clean and noisy clients by measuring the discrepancies in their prediction spaces using the local intrinsic dimension. Clean clients are then leveraged to assist noisy clients by relabeling their candidate noisy samples. Similarly, FedNoRo \cite{fednoro} is a multi-stage method that identifies clean and noisy clients through per-class loss values, after which noisy clients are guided via KD from the global model rather than explicit relabeling. Moreover, FedNoRo adopts DaAgg that differentially weights client contributions based on their estimated noise levels. Building on the noisy client detection of FedNoRo, FedELC \cite{FedELC}
addresses the second stage of label correction by modeling the possible ground-truth label of each noisy sample as a differentiable soft distribution that is jointly optimized with the network parameters.

  \section{Methods}
\label{sec:formatting}
%
We propose a multi-stage FL framework for identifying noisy clients, constructing reliable clean reference representations, and correcting noisy labels.

\begin{figure}[t]
  \centering
  \includegraphics[width=\linewidth]{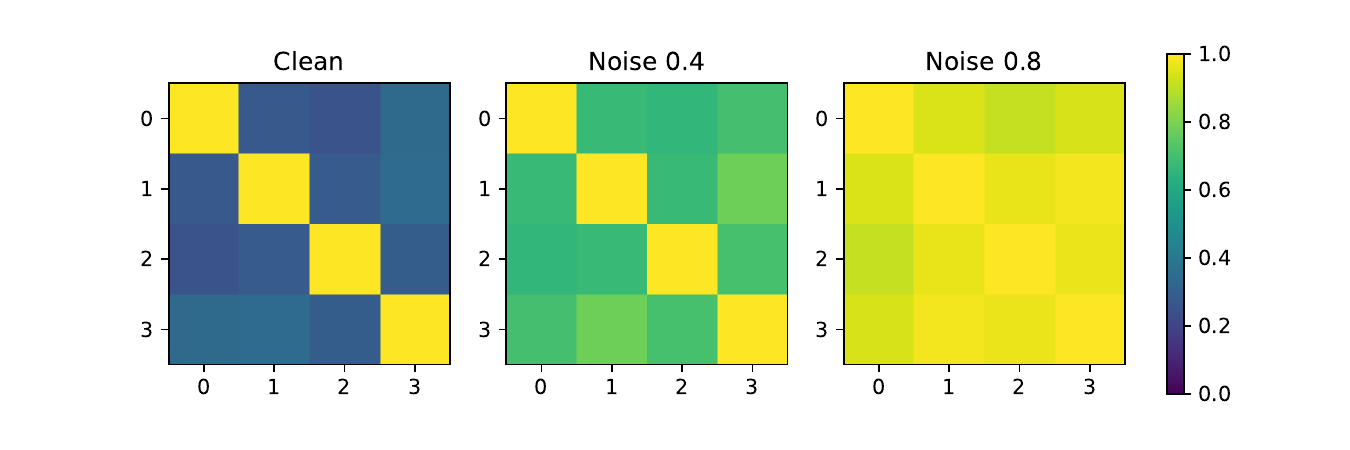}
  \caption{Class subspace similarity matrices for a subset of classes with clean labels (left), $40\%$ symmetric noise (middle), and $80\%$ symmetric noise (right).}\label{fig:example}
\end{figure}

Our approach is motivated by the spectral structure of clean and noisy clients. Under an early-stage trained model, samples from the same class tend to form well-separated clusters in feature space. As a result, clean clients exhibit consistent class-wise spectral structures, characterized by high within-class similarity and low between-class similarity. In contrast, label noise mixes samples from different classes and disrupts this spectral separation. This phenomenon is illustrated in Figure \ref{fig:example}.

We leverage these differences in spectral structure to identify clean and noisy clients. Clean clients are then used to construct an initial global model, while the class-wise spectral structure is further utilized to relabel samples from noisy clients. FL then continues using KD to progressively refine the model.

A key advantage of our formulation is that it eliminates the need for centralized clean data. Instead, it extracts supervisory signals from the internal spectral structure of client feature representations.
This property is particularly desirable in FL, where both raw data sharing and direct inspection of label noise are infeasible. Moreover, our framework does not require estimating a noise transition matrix, which is typically difficult to obtain in federated settings due to heterogeneous data distributions across clients.  

\subsection{Overview}

Let there be \(K\) clients and \(C\) classes. 
Each client \(k\) holds a local dataset
\[
\mathcal{D}_k = \{(x_i, \tilde{y}_i)\}_{i=1}^{N_k},
\]
where \(\tilde{y}_i\) may be corrupted, and $|\mathcal{D}_k|=N_k$.
Our method consists of three stages:

\begin{enumerate}
    \item \textbf{Stage I: Client Identification.} Detect clean and noisy clients from the spectral structure of their class-wise feature matrices, characterized by their top-$r$ singular vectors. 

    \item \textbf{Stage II: Spectral Relabeling of Noisy clients.} Use class subspaces extracted from clean clients to infer corrected labels for samples from noisy clients.

    \item \textbf{Stage III: Noise-Aware Federated Optimization.} Update clean clients with LA supervision, train noisy clients using a hybrid relabeling-and-KD loss, and aggregate all clients using DaAgg.
\end{enumerate}

We first perform a coarse client-level separation to identify a subset of clean clients. 

Once clean clients are identified, their learned representations serve as stable anchors for subsequent sample-level correction.

Figure~\ref{fig:pipeline} illustrates the overall pipeline of our approach.

\subsection{Spectral View of Class Structure}
Given a feature extractor $f_{\boldsymbol{\phi}}$, each sample $\mathbf{x}_i$ is mapped to a feature representation
\[\mathbf{z}_i = f_{\boldsymbol{\phi}}(\mathbf{x}_i) \in \mathbb{R}^d,\]
where $d$ is the feature dimension. For client \( k \) and class \( c \), we construct a class-wise feature matrix 
\[\mathbf{Z}_{k,c} = [\mathbf{z}_1, \mathbf{z}_2, \dots, \mathbf{z}_{n_{k,c}}]^\top \in \mathbb{R}^{n_{k,c} \times d},\]
where $n_{k,c}$ denotes the number of samples labeled as class $c$, i.e., $\tilde{y}_i=c$, at client $k$. 

We then perform singular value decomposition (SVD) on $\mathbf{Z}_{k,c}$:
\begin{equation}
{\mathbf{Z}_{k,c} }= \mathbf{U}_{k,c} \mathbf{\Sigma}_{k,c} \mathbf{V}_{k,c}^{\top},
\end{equation}
and use the leading right singular vector \(\mathbf{v}_{k,c}\) to characterize the spectral structure of class \(c\) at client \(k\). This vector captures the dominant direction of variation for that class. 
If a class is internally consistent, its features tend to align around a relatively stable direction in the feature space. When labels are corrupted, samples from other classes are mixed into the same set, causing the resulting direction to align with multiple class directions rather than a single one \cite{fine}.

To quantify how strongly class structures overlap within a client, we compare their principal directions. 
Using the leading right singular vector of each class, we define the class similarity 
\begin{equation}
[\mathbf{S}_k]_{c,c'} = \left|\mathbf{v}_{k,c}^{\top} \mathbf{v}_{k,c'}\right|. 
\end{equation}
This yields a class similarity matrix \(\mathbf{S}_k \in \mathbb{R}^{C\times C}\) for each client and reflects the principal-angle relationships between pairwise class subspaces.
Clean clients typically exhibit lower off-diagonal similarity, as different classes occupy more separated directions in feature space. Noisy clients, in contrast, exhibit higher off-diagonal similarity because mislabeled samples introduce cross-class mixing in the feature representations. As the noise level increases, this effect becomes more pronounced, leading to progressively larger off-diagonal similarities in $\mathbf{S}_k$. Importantly, this computation is fully local and requires no auxiliary information from other clients or the central server.

\subsection{Stage I: Client Identification}
We use these class-wise spectral signatures to identify clean and noisy clients. 

For each client \(k\), we form a class-similarity matrix \(\mathbf{S}_k\), where entries are computed only for class pairs observed on that client. 
Let \(\mathcal{C}_k = \{c : n_{k,c} > 0\}\) denote the set of classes available at client \(k\). 
We summarize the off-diagonal structure of \(\mathbf{S}_k\) using
\begin{equation} \label{eq:mu_k}
\mu_k =
\frac{1}{|\mathcal{C}_k|(|\mathcal{C}_k|-1)}
\sum_{\substack{c,c' \in \mathcal{C}_k \\ c \ne c'}}
[\mathbf{S}_k]_{c,c'},
\end{equation}
and
\begin{equation} \label{eq:e_k}
e_k =
\frac{1}{|\mathcal{C}_k|(|\mathcal{C}_k|-1)}
\sum_{\substack{c,c' \in \mathcal{C}_k \\ c \ne c'}}
[\mathbf{S}_k]_{c,c'}^2,
\end{equation}
which measure the average off-diagonal similarity and its squared energy, respectively. These statistics quantify the degree of cross-class mixing within each client. For classes absent from client \(k\), the corresponding entries are undefined and excluded from these statistics. Clients with fewer than two observed classes do not yield valid off-diagonal statistics and are excluded from the spectral identification step. We then fit a two-component Gaussian mixture model (GMM) over these descriptors \((\mu_k, e_k)\) and partition clients into two groups: a \emph{clean set} and a \emph{noisy set}.

\subsection{Stage II: Spectral Relabeling for Noisy Clients}

After identifying clean clients in Stage~I, we use their class-wise spectral directions to construct reference subspaces for label correction.
Let $\mathcal{K}_{\text{clean}}$ denote the set of clean clients. We assume that clients in $\mathcal{K}_{\text{clean}}$ collectively provide sufficiently reliable spectral directions across classes, allowing them to serve as class references. 

Specifically, we construct a compact representation of each class $c$, through its dominant direction $\mathbf{v}_c^{(r)}$, and its residual subspace $\mathbf{V}_c^{(n)}$. Potentially corrupted samples are associated with the class whose dominant direction yields the highest alignment, while the corresponding residual subspace produces the lowest projection energy. The client-side spectral extraction procedure is summarized in Algorithm~\ref{alg:extract_spectral}.

\begin{equation}\label{eq:pro_mean_r}
\mathbf{M}^{(r)}_c =
\frac{1}{W_c}
\sum_{k \in \mathcal{K}_\mathrm{clean}}
w_{k,c}\,\mathbf{v}_{k,c}\mathbf{v}_{k,c}^{\top},
\quad
W_c = \sum_{k \in \mathcal{K}_\mathrm{clean}} w_{k,c},
\end{equation}
where \(\mathbf{v}_{k,c}\) is the dominant class direction from client \(k\) and \(w_{k,c}\) is a reliability weight determined by the class sample
count.
The aggregated class direction \(\bar{\mathbf{v}}^{(r)}_c\) is the principal eigenvector of \(\mathbf{M}^{(r)}_c\).

Similarly, for residual subspaces across clean clients we set
\begin{equation} \label{eq:pro_mean_n}
\mathbf{M}^{(n)}_c =
\frac{1}{W_c}
\sum_{k \in \mathcal{K}_\mathrm{clean}}
w_{k,c}\,\mathbf{V}^{(n)\top}_{k,c}\mathbf{V}^{(n)}_{k,c},
\end{equation}
where \(\mathbf{V}^{(n)}_{k,c}\) contains residual directions orthogonal to \(\mathbf{v}_{k,c}\).
The consensus residual subspace \(\bar{\mathbf{V}}^{(n)}_c\) is given by the top \(L\) eigenvectors of \(\mathbf{M}^{(n)}_c\).

Given these spectral references, we evaluate two complementary scores for each sample with feature \(\mathbf{z}_i\):
\begin{align}
S^{(r)}(i,c) &= \left| \mathbf{z}_i^\top \bar{\mathbf{v}}^{(r)}_c \right|, \\
S^{(n)}(i,c) &= \frac{1}{\sqrt{L}}
\left\| \mathbf{z}_i^\top \bar{\mathbf{V}}^{(n)}_c \right\|_2 .
\end{align}

Finally, we derive two class predictions:

\begin{align}
\hat{y}^{(r)}_i &= \arg\max_c S^{(r)}(i,c), \\
\hat{y}^{(n)}_i &= \arg\min_c S^{(n)}(i,c),
\end{align}
and accept a relabel only when
\[
\hat{y}^{(r)}_i = \hat{y}^{(n)}_i .
\]
Thus, the corrected label is
\[
y_i^\star =
\begin{cases}
\hat{y}^{(r)}_i, & \text{if } \hat{y}^{(r)}_i = \hat{y}^{(n)}_i, \\
\tilde{y}_i, & \text{otherwise},
\end{cases}
\]
yielding the updated dataset
\[
\mathcal{D}_k^\star = \{(x_i, y_i^\star)\}_{i=1}^{N_k}.
\]
This agreement rule enforces conservative relabeling by correcting only samples with consistent spectral evidence. The detailed relabeling procedure for a noisy client is given in Algorithm~\ref{alg:relabel}.

\subsection{Noise-aware Federated Optimization}

Once the clean and noisy client sets have been identified, all clients continue to participate in FL, but their local optimization follows a noise-aware strategy. Clean clients, whose labels are considered reliable, are trained with the LA only, whereas noisy clients are trained with a hybrid objective that combines hard targets and teacher guidance. 

\noindent \textbf{Clean clients.}
For each clean client \(k \in \mathcal{K}_{\mathrm{clean}}\), we estimate the empirical class prior from its local labels,
\[
\pi_{k,c} = \frac{n_{k,c}}{\sum_{c'=1}^{C} n_{k,c'}}.
\]
We then define a per-class logit-adjustment:
\[
m_{k,c} = \beta \log(\pi_{k,c} + \epsilon), 
\]
where $\beta > 0$ controls the strength of the adjustment and $\epsilon$ prevents degenerate priors when some classes are absent. If \(f_{\phi_{k}}(x)\) denotes the output logits of the local model, the clean-client objective is
\begin{equation}
\mathcal{L}_{\mathrm{LA}}
=
\mathrm{CE}\!\left(f_{\phi_{k}}(x)+\mathbf{m}_k,\; \tilde{y}\right).
\end{equation}
This follows the same LA formulation used in \cite{fednoro} and \cite{FedELC}. The adjustment compensates for client-level class imbalance, which commonly arises under heterogeneous label distributions.

\noindent \textbf{Noisy clients.}
To leverage global learning, noisy clients are trained using both the hard target \(y_i^\star\) from the relabeling step and the soft predictions from the global model trained on all clients. Let \(f_{\phi_{global}}(\mathbf{x})\) denote the logits produced by the global model. The soft probability distribution for each sample \(\mathbf{x}_i\) is defined as
\begin{equation}
        \mathbf{p}_i = \mathrm{softmax}\!\left(\frac{f_{\phi_{global}}(\mathbf{x}_i)}{\tau}\right),
\end{equation}
where \(\tau\) is the temperature parameter controlling the softness of the distribution.

Before computing the loss, the noisy logits are adjusted using the class-prior correction described earlier. 
Let
\begin{equation}
\mathbf{h}_i = f_{\phi_{k'}}(\mathbf{x}_i) + \mathbf{m}_{k'}
\end{equation}
denote the adjusted logits for noisy client \(k'\). 
The corresponding log-probabilities of the noisy client model are then
\begin{equation}
\mathbf{q}_i = \mathrm{softmax}(\mathbf{h}_i).
\end{equation}
Using these quantities, the noisy-client objective combines KD and hard-label supervision using weight $w_{KD}$ as follows:
\begin{equation}
\mathcal{L}_{\mathrm{LA\text{-}KD}}
=
w_{\mathrm{KD}}\,\mathcal{L}_{\mathrm{KD}}
+
(1-w_{\mathrm{KD}})\,\mathcal{L}_{\mathrm{CE}},
\end{equation}
where
\begin{align}
\mathcal{L}_{\mathrm{KD}} &= 
\mathrm{KL}\!\left(\mathbf{p}_i \,\|\, \mathbf{q}_i\right), \\
\mathcal{L}_{\mathrm{CE}} &= 
\mathrm{NLL}\!\left(\log \mathbf{q}_i,\tilde{y}_i\right).
\end{align}
This formulation allows the noisy to learn from two complementary signals. 
The corrected hard label provides a strong supervisory signal when the spectral relabeling mechanism identifies clear inconsistencies in the original annotation. 
At the same time, the global model offers a softer form of guidance that captures class relationships.
This is particularly helpful when the local label remains uncertain or when the relabeling evidence is weak.

Combining these two sources of supervision makes the optimization more stable in the presence of label noise. 
Hard-label correction addresses samples that are clearly mislabeled, while KD regularizes training dynamics and prevents the model from overfitting to unreliable targets. 
\subsection{Distance-aware Aggregation (DaAgg)}
After local updates, all client models are aggregated into the next global model using a DaAgg rule inspired by RSCFed~\cite{Daagg}. The aggregation starts from the standard sample-size weighting but additionally downweights noisy clients according to how far their local models deviate from those of clean clients. 

Let \(\boldsymbol{\phi}^{(t)}_k\) denote the model parameters of client \(k\)  at round $t$. 
For each noisy client \(k\), we measure its $\ell_2$ distance to the clean set as
\begin{equation}
d_k = \min_{j \in \mathcal{K}_{\mathrm{clean}}}
\left\|\boldsymbol{\phi}^{(t)}_k- \boldsymbol{\phi}^{(t)}_j\right\|_2.
\end{equation}

The minimum distance is used so that each noisy client is evaluated relative to its nearest clean counterpart rather than the clean population average. 
This choice is particularly suitable in heterogeneous settings, where a useful noisy client update may resemble only a subset of the clean clients. Now let \(a_k\) be the normalized sample-size weight of client \(k\). 
The final aggregation weight is defined as
\begin{equation}
\alpha_k
=
\frac{a_k \exp(-d_k)}
{\sum_{j=1}^{K} a_j \exp(-d_j)},
\end{equation}
where \(d_k=0\) for clean clients.
The global model is then updated as
\begin{equation}
\boldsymbol{\phi}^{(t+1)} = \sum_{k=1}^{K} \alpha_k \boldsymbol{\phi}^{(t)}_k.
\end{equation}
This aggregation mechanism introduces an additional level of robustness. 
Even after relabeling and KD, some noisy clients may still produce updates that deviate from the clean population. 
DaAgg reduces their influence without excluding them entirely, allowing useful information to contribute while suppressing unreliable updates.

The complete procedure after Stage~II proceeds as follows. 
At each communication round, the current global model is broadcast to all clients. 
Clean clients update their models using the LA objective, while noisy clients are trained with the hybrid LA-KD objective. During each relabeling step (performed every \(R\) communication rounds), feature representations are extracted by the model trained on clean clients. The spectral information of these features is then used to assist noisy clients in relabeling their potentially corrupted samples. 
Finally, all local models are aggregated using DaAgg to produce the next global model, while the updates from clean clients are separately averaged to obtain the clean reference model used to guide the subsequent relabeling step.
The complete training procedure is summarized in Algorithm~\ref{alg:ours_v2}.

\begin{algorithm}[t]
\caption{FedSIR}
\label{alg:ours_v2}
\begin{algorithmic}[1]
\Require Client datasets \(\{\mathcal{D}_k\}_{k=1}^K\), initial model \(\boldsymbol{\phi}^{(0)}\), communication rounds \(T\), identification epochs \(E_1\), training epochs \(E_2\), \(R\) relabeling period
\Ensure Final global model \(\boldsymbol{\phi}^{(T)}\)

\Statex \textbf{Stage I: Client identification}
\For{each client \(k = 1,\dots,K\)}
    \State Initialize local model \(\boldsymbol{\phi}_k \gets \boldsymbol{\phi}^{(0)}\)
    \State \(
    \boldsymbol{\phi}_k \gets \mathrm{LocalTrain}(\boldsymbol{\phi}_k,\mathcal{D}_k,E_1, \mathcal{L}_{CE})
    \)
    \State  $ \{\mathbf{v}_{k,c}\}\gets$ Alg.~\ref{alg:extract_spectral}
        $(\mathcal{D}_k, \boldsymbol{\phi}_k, 0)$

    \State Compute \((\mu_k,e_k)\) from \(\mathbf{S}_k\) using \eqref{eq:mu_k} and \eqref{eq:e_k}
    \State Send \((\mu_k,e_k)\) and \(\boldsymbol{\phi}_k\) to the server

\EndFor
\State Fit a two-component GMM on \(\{(\mu_k,e_k)\}_{k=1}^K\)
\State Partition clients into \(\mathcal{K}_{\mathrm{clean}}\) and \(\mathcal{K}_{\mathrm{noisy}}\)

\State \(
\boldsymbol{\phi}^{(1)}
\gets
\mathrm{FedAvg}_{k \in \mathcal{K}_{\mathrm{clean}}}(\boldsymbol{\phi}_k)
\)

\Statex \textbf{Stage II--III: Relabeling and Training}
\For{{round \(t = 1,\dots, T\)}}
    \For{each client $k \in \mathcal{K}_{\mathrm{clean}}$}
        \State Initialize \(\boldsymbol{\phi}_k^{\,(t)} \gets \boldsymbol{\phi}^{(t)}\)
        \State \(
        \boldsymbol{\phi}_k^{\,(t)}
        \gets
        \mathrm{LocalTrain}(\boldsymbol{\phi}_k^{\,(t)},\mathcal{D}_k, E_2, \mathcal{L}_{\mathrm{LA}})
        \)
        \If {$t \% R == 0$}
            \State $(\{\mathbf{v}_{k,c}, \mathbf{V}^{(n)}_{k,c}\}) \gets 
            \text{Alg.~\ref{alg:extract_spectral}}(\mathcal{D}_k, \boldsymbol{\phi}_k^{(t)}, L)$

            \State Send \(\boldsymbol{\phi}_k^{\,(t)}\), $\{(\mathbf{v}_{k,c}, \mathbf{V}^{(n)}_{k,c})\}$ to server
        \Else
        \State Send \(\boldsymbol{\phi}_k^{\,(t)}\) to server
        \EndIf
    \EndFor
    \State 
    Apply \eqref{eq:pro_mean_r} and \eqref{eq:pro_mean_n} on \(
    \left(
    \{(\mathbf{v}_{k,c}, \mathbf{V}^{(n)}_{k,c})\}
    \right)\) to get
    \Statex\hspace{0.6cm}\(\bar{\mathbf{v}}^{(r)},\bar{\mathbf{V}}^{(n)}
      \)
    \For{each client $k \in \mathcal{K}_{\mathrm{noisy}}$}
        \If {$t \% R == 0$}
        \State
            \(
            \mathcal{D}_k^\star
            \gets
            \text{Alg.~\ref{alg:relabel}}\!\left(
            \mathcal{D}_k,
            \boldsymbol{\phi}_{\mathrm{clean}}^{(t)},
            \bar{\mathbf{v}}^{(r)},\bar{\mathbf{V}}^{(n)}            
            \right)\)
        \EndIf
        \State 
        \(
        \boldsymbol{\phi}_k^{\,(t)}
        \gets
        \mathrm{LocalTrain}(\boldsymbol{\phi}_k^{\,(t)},\mathcal{D}_k^\star,E_2, \mathcal{L}_{\mathrm{LA\text{-}KD}})\)

    \EndFor
    \State \(
    \boldsymbol{\phi}^{(t+1)}
    \gets
    \mathrm{DaAgg}\!\left(
    \{\boldsymbol{\phi}_k^{\,(t)}\}_{k=1}^K,
    \mathcal{K}_{\mathrm{clean}},
    \mathcal{K}_{\mathrm{noisy}}
    \right) 
    \)
    \State \(
    \boldsymbol{\phi}_{\mathrm{clean}}^{(t+1)}
    \gets
    \mathrm{FedAvg}_{k \in \mathcal{K}_{\mathrm{clean}}}
    (\boldsymbol{\phi}_k^{\,(t)})
    \)
   
\EndFor

\State \Return \(\boldsymbol{\phi}^{(T)}\)
\end{algorithmic}
\end{algorithm}

\begin{algorithm}[t]
\caption{Client Spectral Structure Extraction}
\label{alg:extract_spectral}
\begin{algorithmic}[1]

\Function{\textsc{ExtractSpectral}}{$\mathcal{D}_k,\boldsymbol{\phi}_k,L$}
    \State Extract features $\mathbf{z}_i = f_{\boldsymbol{\phi}_k}(\mathbf{x}_i)$ for all $(\mathbf{x}_i,\tilde{y}_i) \in \mathcal{D}_k$
    \State Define observed class set $\mathcal{C}_k \gets \{c : n_{k,c} > 0\}$
    \For{each class $c \in \mathcal{C}_k$}
        \State Construct class-wise feature matrix $\mathbf{Z}_{k,c}$
        \Statex \hspace{\algorithmicindent}\quad \ \ from samples with $\tilde{y}_i = c$
        \State Compute SVD $\mathbf{Z}_{k,c} = \mathbf{U}_{k,c}\mathbf{\Sigma}_{k,c}\mathbf{V}_{k,c}^{\top}$
        \State $\mathbf{v}_{k,c} \gets$ leading right singular vector of $\mathbf{Z}_{k,c}$
        \If{$L > 0$}
        \State $\mathbf{V}^{(n)}_{k,c} \gets$ top $L$ residual right singular
        \Statex \hspace{\algorithmicindent}\quad\quad\quad vectors of $\mathbf{Z}_{k,c}$
        \EndIf
    \EndFor
    \If{$L > 0$}
        \State \Return $\{(\mathbf{v}_{k,c}, \mathbf{V}^{(n)}_{k,c})\}_{c \in \mathcal{C}_k}$
    \Else
        \State \Return $\{\mathbf{v}_{k,c}\}_{c \in \mathcal{C}_k}$
    \EndIf
\EndFunction

\end{algorithmic}
\end{algorithm}

\begin{algorithm}[t]
\caption{Spectral Relabeling for a Noisy Client}
\label{alg:relabel}
\begin{algorithmic}[1]

\Function{\textsc{Relabel}}{$\mathcal{D}_k,\boldsymbol{\phi}_{\mathrm{clean}},\{\bar{\mathbf{v}}^{(r)}_c,\bar{\mathbf{V}}^{(n)}_c\}_{c=1}^C$}

\State $\mathcal{D}_k^\star \gets \emptyset$
    \For{each $(\mathbf{x}_i,\tilde y_i)\in\mathcal D_k$}
    \State $\mathbf{z}_i \gets f_{\boldsymbol{\phi}_{\mathrm{clean}}}(\mathbf{x}_i)$
        \For{each class $c=1,\dots,C$}
            \State $
            S^{(r)}(i,c)
            =
            \left|
            \mathbf{z}_i^\top \bar{\mathbf{v}}^{(r)}_c
            \right|
            $
            \State $
            S^{(n)}(i,c)
            =
            \frac{1}{\sqrt{L}}
            \left\|
            \mathbf{z}_i^\top \bar{\mathbf{V}}^{(n)}_c
            \right\|_2
            $
        \EndFor
        \State $\hat{y}^{(r)}_i \gets \arg\max_c S^{(r)}(i,c)$
        \State $\hat{y}^{(n)}_i \gets \arg\min_c S^{(n)}(i,c)$
        \If{$\hat{y}^{(r)}_i = \hat{y}^{(n)}_i$}
            \State $y_i^\star \gets \hat{y}^{(r)}_i$
        \Else
            \State $y_i^\star \gets \tilde{y}_i$
        \EndIf
        \State $\mathcal{D}_k^\star \gets \mathcal{D}_k^\star \cup \{(\mathbf{x}_i,y_i^\star)\}$
    \EndFor
    \State \Return $\mathcal{D}_k^\star$
\EndFunction

\end{algorithmic}
\end{algorithm}
\section{Results}

We evaluate the proposed method on CIFAR-10 under symmetric label noise with 10 federated clients. 
To simulate data heterogeneity, the training data samples are partitioned across clients using a Dirichlet distribution with concentration parameters $\alpha \in \{0.1, 0.5, 2\}$, where smaller values of $\alpha$ correspond to stronger non-IID data distributions. 
To introduce heterogeneity in client-level noise, the number of clean clients varies across different settings. Specifically, the number of clean clients is $3$, $3$, and $5$ for $\alpha = 0.1$, $0.5$, and $2$, respectively. Furthermore, we evaluate multiple noise rates ranging from $30\%$ to $90\%$ to assess the robustness of the proposed framework under different levels of label corruption.


All experiments are conducted for $T = 100$ communication rounds. 
In Stage~I, each client performs local training for $E_1=5$ epochs for client identification, using a learning rate (LR) of $5\times10^{-5}$ and a weight decay (WD) of $2\times10^{-2}$. In Stage~II, local updates are performed for $E_2=1$ epoch per round using LR $=3\times10^{-4}$ and WD $= 5\times10^{-4}$, while spectral relabeling is performed every $R=20$ communication rounds using the top $L = 12$ residual right singular vectors. In all experiments, local models are optimized using the Adam optimizer. We use an ImageNet-pretrained ResNet-18 \cite{He_2016_CVPR} as the backbone model. 
We compare our method with standard FL baselines and recent noise-robust FL approaches. In addition, we include a pruning baseline that evaluates the performance obtained by excluding identified noisy clients.  

Table~\ref{tab:cifar10} shows that our method consistently achieves the best performance across nearly all noise levels and heterogeneity settings. 
In the highly non-IID case ($\alpha=0.1$), our method remains robust even under severe noise, outperforming baselines such as FedELC, FedNoRo, and FedNed. 
As the data distribution becomes less heterogeneous ($\alpha=0.5$ and $\alpha=2$), the performance of all methods generally improves; however, our approach continues to maintain a clear advantage, particularly at high noise rates where reliable client identification and conservative relabeling become more critical. 
These results indicate that the proposed spectral client identification and relabeling strategy provides robustness to both label noise and client heterogeneity.

\begin{table*}[t]
\centering
\caption{Results on CIFAR-10 under symmetric label noise. \textbf{Bold} indicates the best result and \underline{underline} indicates the second-best result.
}
\label{tab:cifar10}
\begin{adjustbox}{width=\textwidth}
\begin{tabular}{c c l c c c c c c c}
\toprule
\multicolumn{3}{c}{} & \multicolumn{7}{c}{Symmetric Noise} \\
\cmidrule(lr){4-10}
$\alpha$ & \# Clean Clients & Method 
& 30\% & 40\% & 50\% & 60\% & 70\% & 80\% & 90\% \\
\midrule

\multirow{10}{*}{0.1} & \multirow{10}{*}{5}
& FedAvg \cite{mcmahan} & 73.67$\pm$0.89 & 73.62$\pm$0.83 & 71.92$\pm$0.83 & 73.60$\pm$0.84 & 73.21$\pm$0.87 & 70.39$\pm$0.94 & 68.90$\pm$0.97 \\
& & FedProx \cite{fedprox} & 73.14$\pm$0.86 & 72.53$\pm$0.87 & 71.60$\pm$0.90 & 72.65$\pm$0.88 & 72.11$\pm$0.86 & 71.24$\pm$0.87 & 68.72$\pm$0.90 \\
& & RoFL \cite{rofl}& 43.73$\pm$1.01 & 44.13$\pm$0.99 & 44.45$\pm$1.00 & 42.81$\pm$0.98 & 35.76$\pm$0.92 & 43.96$\pm$0.98 & 40.75$\pm$0.96 \\
& & RHFL \cite{RHFL}& 27.72$\pm$5.88 & 27.51$\pm$6.20 & 26.61$\pm$5.94 & 24.98$\pm$6.10 & 25.01$\pm$5.97 & 22.37$\pm$6.63 & 20.56$\pm$7.56 \\
& & FedLSR \cite{fedlsr}& 64.82$\pm$0.99 & 63.99$\pm$0.90 & 61.81$\pm$0.96 & 63.70$\pm$0.96 & 61.38$\pm$0.98 & 63.43$\pm$0.98 & 61.68$\pm$0.94 \\
& & FedCorr \cite{fedcorr}& 62.46$\pm$1.00 & 60.62$\pm$0.91 & 57.28$\pm$0.96 & 55.12$\pm$1.01 & 54.49$\pm$0.98 & 53.46$\pm$0.97 & 52.02$\pm$0.99 \\
& & FedNed \cite{fedned}& 74.16$\pm$0.86 & 72.44$\pm$0.90 & 74.60$\pm$0.86 & 74.11$\pm$0.85 & 73.29$\pm$0.87 & 73.83$\pm$0.86 & 73.12$\pm$0.87 \\
& & FedELC \cite{FedELC}& \underline{77.71$\pm$0.84} & \underline{78.18$\pm$0.80} & \underline{78.07$\pm$0.81} & \underline{77.88$\pm$0.80} & \underline{77.29$\pm$0.81} & \underline{78.43$\pm$0.82} & \underline{77.72$\pm$0.82} \\
& & FedNoRo \cite{fednoro}& 77.65$\pm$0.76 & 76.56$\pm$0.84 & 76.47$\pm$0.80 & 76.73$\pm$0.81 & 76.80$\pm$0.81 & 76.83$\pm$0.83 & 76.11$\pm$0.86 \\
& & \textbf{FedSIR (Ours)} & \textbf{78.65$\pm$0.81} & \textbf{78.42$\pm$0.83} & \textbf{78.10$\pm$0.78} & \textbf{78.31$\pm$0.80} & \textbf{77.93$\pm$0.81} & \textbf{77.93$\pm$0.81} & \textbf{77.90$\pm$0.77} \\
\cmidrule(lr){2-10}
& 5 & Pruning & \multicolumn{7}{c}{76.91$\pm$0.82} \\
& 10 & FedAvg & \multicolumn{7}{c}{76.34$\pm$0.82} \\
\midrule

\multirow{10}{*}{0.5} & \multirow{10}{*}{3}
& FedAvg \cite{mcmahan}& 83.14$\pm$0.74 & 81.56$\pm$0.78 & 80.78$\pm$0.78 & 77.35$\pm$0.83 & 71.63$\pm$0.89 & 62.42$\pm$0.92 & 40.24$\pm$0.97 \\
& & FedProx \cite{fedprox}& 82.21$\pm$0.77 & 80.64$\pm$0.78 & 79.29$\pm$0.81 & 77.25$\pm$0.84 & 75.88$\pm$0.84 & 64.53$\pm$0.94 & 37.64$\pm$0.94 \\
& & RoFL \cite{rofl}& 71.18$\pm$0.84 & 67.27$\pm$0.94 & 59.28$\pm$0.93 & 59.82$\pm$0.94 & 61.60$\pm$1.03 & 54.98$\pm$0.95 & 56.46$\pm$0.99 \\
& & RHFL \cite{RHFL}& 48.37$\pm$2.47 & 44.91$\pm$4.06 & 41.95$\pm$4.59 & 40.07$\pm$4.55 & 36.34$\pm$5.62 & 27.62$\pm$9.20 & 22.34$\pm$12.06 \\
& & FedLSR \cite{fedlsr}& 71.67$\pm$0.89 & 68.53$\pm$0.89 & 69.19$\pm$0.89 & 67.06$\pm$0.91 & 63.54$\pm$0.94 & 59.06$\pm$0.94 & 51.94$\pm$0.90 \\
& & FedCorr \cite{fedcorr}& 75.23$\pm$0.83 & 73.48$\pm$0.84 & 72.95$\pm$0.87 & 71.73$\pm$0.85 & 67.80$\pm$0.94 & 65.58$\pm$0.93 & 64.37$\pm$0.96 \\
& & FedNed \cite{fedned}& 77.16$\pm$0.85 & 77.87$\pm$0.81 & 77.25$\pm$0.80 & 77.22$\pm$0.83 & 76.95$\pm$0.81 & 76.78$\pm$0.81 & 76.59$\pm$0.83 \\
& & FedELC \cite{FedELC}& \underline{82.85$\pm$0.78} & \underline{82.45$\pm$0.71} & \underline{82.84$\pm$0.73} & \underline{82.28$\pm$0.75} & \underline{82.12$\pm$0.76} & \underline{81.86$\pm$0.79} & 77.74$\pm$0.78 \\
& & FedNoRo \cite{fednoro}& 82.65$\pm$0.75 & 81.86$\pm$0.76 & 81.73$\pm$0.75 & 81.55$\pm$0.72 & 81.43$\pm$0.72 & 80.86$\pm$0.80 & \underline{81.41$\pm$0.77} \\
& & \textbf{FedSIR (Ours)} & \textbf{84.13$\pm$0.75} & \textbf{83.96$\pm$0.71} & \textbf{83.94$\pm$0.72} & \textbf{83.55$\pm$0.70} & \textbf{83.45$\pm$0.70} & \textbf{83.56$\pm$0.68} & \textbf{83.15$\pm$0.74} \\
\cmidrule(lr){2-10}
& 3 & Pruning & \multicolumn{7}{c}{78.70$\pm$0.79} \\
& 10 & FedAvg & \multicolumn{7}{c}{86.81$\pm$0.61} \\
\midrule
\multirow{10}{*}{2} & \multirow{10}{*}{3}
& FedAvg \cite{mcmahan} & 84.61$\pm$0.68 & 83.73$\pm$0.72 & 82.23$\pm$0.76 & 81.67$\pm$0.78 & 78.94$\pm$0.83 & 73.55$\pm$0.86 & 40.90$\pm$0.98 \\
& & FedProx \cite{fedprox}& 83.98$\pm$0.73 & 82.70$\pm$0.73 & 81.06$\pm$0.77 & 80.31$\pm$0.80 & 78.73$\pm$0.79 & 72.94$\pm$0.86 & 39.24$\pm$1.00 \\
& & RoFL \cite{rofl}& 78.51$\pm$0.81 & 77.69$\pm$0.81 & 74.69$\pm$0.82 & 73.56$\pm$0.87 & 71.83$\pm$0.87 & 69.36$\pm$0.90 & 69.39$\pm$0.86 \\
& & RHFL \cite{RHFL}& 63.28$\pm$2.25 & 61.15$\pm$2.12 & 58.18$\pm$3.37 & 55.66$\pm$4.04 & 49.81$\pm$6.08 & 35.47$\pm$11.06 & 25.82$\pm$15.07 \\
& & FedLSR \cite{fedlsr}& 73.68$\pm$0.84 & 73.87$\pm$0.89 & 72.43$\pm$0.91 & 70.27$\pm$0.90 & 68.63$\pm$0.89 & 64.64$\pm$0.88 & 56.12$\pm$1.01 \\
& & FedCorr \cite{fedcorr}& 78.48$\pm$0.78 & 77.34$\pm$0.82 & 76.81$\pm$0.83 & 76.65$\pm$0.84 & 75.84$\pm$0.84 & 75.46$\pm$0.87 & 74.97$\pm$0.87 \\
& & FedNed \cite{fedned}& 82.27$\pm$0.73 & 81.95$\pm$0.77 & 81.15$\pm$0.76 & 81.78$\pm$0.77 & 81.25$\pm$0.75 & 81.44$\pm$0.76 & 81.00$\pm$0.76 \\
& & FedELC \cite{FedELC}& 84.00$\pm$0.70 & 83.54$\pm$0.75 & 83.62$\pm$0.71 & 83.11$\pm$0.75 & 82.94$\pm$0.73 & 82.42$\pm$0.73 & 79.73$\pm$0.78 \\
& & FedNoRo \cite{fednoro}& \underline{84.93$\pm$0.70} & \underline{84.51$\pm$0.72} & \underline{83.97$\pm$0.72} & \underline{83.80$\pm$0.73} & \underline{83.26$\pm$0.75} & \underline{82.50$\pm$0.79} & \underline{82.59$\pm$0.74} \\
& & \textbf{FedSIR (Ours)} & \textbf{85.72$\pm$0.70} & \textbf{85.23$\pm$0.69} & \textbf{85.19$\pm$0.70} & \textbf{84.95$\pm$0.67} & \textbf{84.76$\pm$0.72} & \textbf{84.26$\pm$0.71} & \textbf{84.40$\pm$0.67} \\

\cmidrule(lr){2-10}
& 3 & Pruning & \multicolumn{7}{c}{82.23$\pm$0.73} \\
& 10 & FedAvg & \multicolumn{7}{c}{88.27$\pm$0.68} \\
\bottomrule
\end{tabular}
\end{adjustbox}
\end{table*}

\section{Conclusion}
We introduced \textbf{FedSIR}, a robust FL framework for noisy-label settings that exploits the consistency of class-wise feature representations to identify clean and noisy clients. The identified clean clients are then used to construct spectral references that enable noisy clients to relabel potentially corrupted samples by associating them with the closest clean class subspace. Furthermore, our framework integrates LA, KD, and DaAgg to stabilize federated optimization.  

Results across a range of label-noise levels and non-IID data distributions show that \textbf{FedSIR} consistently outperforms state-of-the-art methods, highlighting the effectiveness of spectral information as a reliable supervisory signal in noisy-label FL. Overall, our findings indicate that client-level spectral geometry provides a principled foundation for robust FL under corrupted supervision. 

An important direction for future work is to extend the proposed framework to handle more complex noise patterns, such as asymmetric label noise.

\section{Acknowledgement}
The work of S.~Gholami, T.~Haghighi, and M.~N.~Alam was supported by the U.S. National Institute of Health - National Eye Institute under Grants R15EY035804 and R21EY035271. The work of A.~Ali and A.~Arafa was supported by the U.S. National Science Foundation under Grant ECCS 21-46099.

{
    \small
    \bibliographystyle{ieeenat_fullname}
    \bibliography{main}
}

\newpage

\clearpage
\setcounter{page}{1}
\maketitlesupplementary
\section{Ablation Study}
To better understand the contribution of each component of FedSIR, we perform an ablation study under symmetric label noise with Dirichlet heterogeneity parameter $\alpha=1$. In this setting, three clients are identified as clean and used to construct the spectral reference model. We evaluate several variants of the proposed framework by removing one component at a time: the spectral relabeling mechanism, LA, KD, and DaAgg. The results are summarized in Table~\ref{tab:ablation}.

The full FedSIR model consistently achieves the best performance across noise levels exceeding 50\%. Removing the spectral relabeling component leads to the most degradation, particularly under high noise rates. For example, at $90\%$ noise, the accuracy drops from $84.51\%$ to $80.00\%$, indicating that the relabeling mechanism plays a critical role in correcting corrupted supervision on noisy clients.

LA mainly compensates for class imbalance across clients. Removing this component results in a moderate but consistent decrease in accuracy, confirming its significance for stabilizing local optimization under heterogeneous label distributions.

KD also contributes to performance stability by providing soft guidance from the global model. When KD is removed, performance decreases slightly across most noise levels, suggesting that KD helps regularize the training of noisy clients and prevents overfitting to unreliable labels.

Finally, removing the DaAgg step slightly improves performance at lower noise rates but degrades robustness at higher noise levels. This behavior suggests that DaAgg primarily acts as a safeguard against highly corrupted client updates rather than improving average-case performance when noise is moderate.

Overall, the ablation results demonstrate that each component contributes to the robustness of the proposed framework, with spectral relabeling playing the critical role in handling severe label corruption.

\begin{table*}[t] \centering \caption{Ablation study of FedSIR on CIFAR-10 with symmetric label noise and Dirichlet parameter $\alpha=1$. Removing each component shows its contribution to the overall framework. \textbf{Bold} indicates the best result and \underline{underline} indicates the second-best result.} \label{tab:ablation} \begin{adjustbox}{width=\textwidth} \begin{tabular}{c c l c c c c c c c} \toprule \multicolumn{3}{c}{} & \multicolumn{7}{c}{Symmetric Noise} \\ \cmidrule(lr){4-10} $\alpha$ & \# Clean Clients & Method & 30\% & 40\% & 50\% & 60\% & 70\% & 80\% & 90\% \\ \midrule \multirow{5}{*}{1} & \multirow{5}{*}{3} & w/o relabeling & 85.13$\pm$0.73 & 84.81$\pm$0.73 & 84.29$\pm$0.70 & 83.90$\pm$0.76 & 82.85$\pm$0.76 & 82.06$\pm$0.75 & 80.00$\pm$0.77 \\ & & w/o LA & 84.93$\pm$0.69 & 84.60$\pm$0.76 & 84.63$\pm$0.70 & 84.38$\pm$0.75 & 84.15$\pm$0.75 & 84.04$\pm$0.71 & 84.14$\pm$0.72 \\ & & w/o KD & \underline{85.49$\pm$0.69} & 85.16$\pm$0.68 & \underline{85.03$\pm$0.70} & \underline{84.63$\pm$0.66} & \underline{84.43$\pm$0.69} & \underline{84.16$\pm$0.73} & \underline{84.37$\pm$0.68} \\ & & w/o DaAgg & \textbf{85.65$\pm$0.70} & \textbf{85.33$\pm$0.73} & \textbf{85.25$\pm$0.68} & 84.41$\pm$0.70 & 83.90$\pm$0.71 & 83.95$\pm$0.73 & 83.73$\pm$0.72 \\ && \textbf{Ours} & 85.21$\pm$0.70 & \underline{85.28$\pm$0.68} & 84.86$\pm$0.70 & \textbf{84.68$\pm$0.71} & \textbf{84.54$\pm$0.73} & \textbf{84.35$\pm$0.74} & \textbf{84.51$\pm$0.72} \\ \midrule & 3 & Pruning & \multicolumn{7}{c}{81.78$\pm$0.80} \\ & 10 & FedAvg & \multicolumn{7}{c}{87.23$\pm$0.66} \\ \bottomrule \end{tabular} \end{adjustbox} \end{table*}

\section{Results on CIFAR-100}

We further evaluate our method on CIFAR-100 under symmetric label noise in a federated setting with 10 clients. Compared with CIFAR-10, CIFAR-100 contains a significantly larger number of classes, which makes the learning problem more challenging under both label noise and non-IID data. In particular, under strong non-IID settings (small $\alpha$), each client tends to contain few classes, requiring more clean clients to ensure sufficient class coverage for reliable spectral reference construction. 
To simulate non-IID data, this dataset is partitioned using a Dirichlet distribution with $\alpha \in \{0.3, 0.5, 2\}$.

Table~\ref{tab:cifar100} reports the classification accuracy under symmetric noise rates ranging from 30\% to 90\%. Despite the increased difficulty of CIFAR-100, our method consistently achieves the best performance across all noise levels and heterogeneity settings. In particular, our approach maintains stable performance even under extremely high noise levels (e.g., 80\%–90\%).

These results further demonstrate that the proposed spectral client identification and relabeling strategy scales well to more challenging multi-class settings and remains robust to both severe label noise and client heterogeneity.

\begin{table*}[b]
\centering
\caption{Results on CIFAR-100 under symmetric label noise. \textbf{Bold} indicates the best result and \underline{underline} indicates the second-best result.}
\label{tab:cifar100}
\begin{adjustbox}{width=\textwidth}
\begin{tabular}{c c l c c c c c c c}
\toprule
\multicolumn{3}{c}{} & \multicolumn{7}{c}{Symmetric Noise} \\
\cmidrule(lr){4-10}
$\alpha$ & \# Clean Clients & Method
& 30\% & 40\% & 50\% & 60\% & 70\% & 80\% & 90\% \\
\midrule

\multirow{10}{*}{0.3} & \multirow{10}{*}{7}
& FedAvg \cite{mcmahan} & 57.36$\pm$0.96 & 56.92$\pm$0.92 & 55.93$\pm$0.94 & 55.60$\pm$0.95 & 55.02$\pm$0.99 & 54.07$\pm$0.97 & 53.10$\pm$1.00 \\
& & FedProx \cite{fedprox}& 57.55$\pm$0.98 & 57.05$\pm$1.02 & 56.89$\pm$1.01 & 55.80$\pm$0.95 & 55.66$\pm$0.97 & 54.88$\pm$0.95 & 54.27$\pm$1.00 \\
& & RoFL \cite{rofl}& 46.77$\pm$0.99 & 45.75$\pm$0.96 & 46.03$\pm$0.98 & 42.82$\pm$0.97 & 41.46$\pm$0.95 & 42.22$\pm$0.98 & 40.72$\pm$0.94 \\
& & RHFL \cite{RHFL}& 18.64$\pm$2.01 & 18.53$\pm$2.05 & 17.76$\pm$2.67 & 17.25$\pm$2.50 & 16.86$\pm$3.55 & 15.51$\pm$4.90 & 13.37$\pm$4.94 \\
& & FedLSR \cite{fedlsr}& 47.55$\pm$1.06 & 46.99$\pm$1.01 & 46.60$\pm$0.99 & 46.14$\pm$0.99 & 44.73$\pm$1.00 & 44.74$\pm$1.03 & 43.53$\pm$0.98 \\
& & FedCorr \cite{fedcorr}& 50.72$\pm$1.04 & 50.56$\pm$1.02 & 50.23$\pm$1.00 & 48.71$\pm$0.94 & 47.77$\pm$0.95 & 47.11$\pm$0.96 & 47.24$\pm$0.99 \\
& & FedNed \cite{fedned}& 56.18$\pm$0.92 & 56.46$\pm$0.95 & 56.56$\pm$1.03 & 56.23$\pm$1.01 & 56.58$\pm$0.96 & 56.21$\pm$0.98 & 56.36$\pm$0.97 \\
& & FedELC\cite{FedELC}& \underline{58.65$\pm$0.93} & \underline{58.45$\pm$0.95} & \underline{58.56$\pm$0.98} & \underline{58.14$\pm$0.95} & \underline{58.38$\pm$0.97} & \underline{58.17$\pm$0.94} & \underline{57.99$\pm$0.93} \\
& & FedNoRo\cite{fednoro}& 56.90$\pm$0.97 & 56.90$\pm$0.94 & 56.74$\pm$0.98 & 56.46$\pm$0.99 & 56.41$\pm$0.97 & 56.51$\pm$0.97 & 55.92$\pm$0.98 \\
& & \textbf{FedSIR (Ours)} & \textbf{59.43$\pm$0.95} & \textbf{59.53$\pm$0.96} & \textbf{58.94$\pm$0.92} & \textbf{59.18$\pm$0.99} & \textbf{58.81$\pm$0.95} & \textbf{58.78$\pm$0.94} & \textbf{58.43$\pm$0.93} \\
\cmidrule(lr){2-10}
& 7 & Pruning & \multicolumn{7}{c}{56.74$\pm$0.95} \\
& 10 & FedAvg & \multicolumn{7}{c}{60.67$\pm$0.93} \\

\midrule

\multirow{10}{*}{0.5} & \multirow{10}{*}{6}
& FedAvg \cite{mcmahan} & 57.38$\pm$1.01 & 57.15$\pm$0.97 & 56.05$\pm$0.99 & 54.95$\pm$0.96 & 54.65$\pm$0.94 & 52.62$\pm$0.93 & 51.25$\pm$1.04 \\
& & FedProx \cite{fedprox}& 57.72$\pm$1.00 & 57.17$\pm$0.96 & 55.95$\pm$0.92 & 55.33$\pm$1.00 & 55.34$\pm$0.97 & 53.58$\pm$0.98 & 52.14$\pm$0.95 \\
& & RoFL \cite{rofl}& 51.20$\pm$0.99 & 48.76$\pm$0.97 & 48.44$\pm$0.96 & 46.12$\pm$0.96 & 47.02$\pm$1.00 & 44.92$\pm$1.01 & 45.24$\pm$0.96 \\
& & RHFL \cite{RHFL}& 21.47$\pm$2.08 & 21.33$\pm$2.00 & 21.08$\pm$2.66 & 19.23$\pm$3.27 & 17.85$\pm$4.20 & 16.32$\pm$5.43 & 15.39$\pm$6.14 \\
& & FedLSR \cite{fedlsr}& 48.50$\pm$0.95 & 47.82$\pm$0.98 & 46.02$\pm$1.00 & 45.53$\pm$1.02 & 44.13$\pm$0.99 & 43.48$\pm$1.01 & 42.38$\pm$1.01 \\
& & FedCorr \cite{fedcorr}& 50.85$\pm$0.96 & 50.80$\pm$0.97 & 50.37$\pm$1.01 & 50.34$\pm$0.94 & 50.03$\pm$0.97 & 49.65$\pm$1.04 & 49.52$\pm$0.95 \\
& & FedNed \cite{fedned}& 56.47$\pm$0.97 & 55.97$\pm$0.91 & 56.04$\pm$1.00 & 55.55$\pm$0.98 & 56.05$\pm$0.94 & 55.70$\pm$0.94 & 55.27$\pm$0.94 \\
& & FedELC\cite{FedELC}& 57.91$\pm$0.97 & 57.85$\pm$0.92 & \underline{57.86$\pm$0.97} & 57.91$\pm$0.99 & 57.64$\pm$0.97 & \underline{57.30$\pm$0.96} & \underline{57.85$\pm$0.95} \\
& & FedNoRo\cite{fednoro}& \underline{58.59$\pm$0.98} & \underline{58.22$\pm$0.95} & 57.74$\pm$1.00 & \underline{57.95$\pm$0.97} & \underline{57.70$\pm$0.97} & 57.10$\pm$0.97 & 57.07$\pm$0.98 \\
& & \textbf{FedSIR (Ours)} & \textbf{59.74$\pm$0.94} & \textbf{59.73$\pm$0.91} & \textbf{59.60$\pm$0.98} & \textbf{59.30$\pm$0.94} & \textbf{59.08$\pm$0.96} & \textbf{58.88$\pm$0.95} & \textbf{58.56$\pm$0.94} \\
\cmidrule(lr){2-10}
& 6 & Pruning & \multicolumn{7}{c}{56.57$\pm$0.93} \\
& 10 & FedAvg & \multicolumn{7}{c}{61.91$\pm$0.93} \\

\midrule

\multirow{10}{*}{2} & \multirow{10}{*}{5}
& FedAvg \cite{mcmahan}& 58.81$\pm$0.96 & 57.88$\pm$0.96 & 56.77$\pm$1.02 & 55.90$\pm$0.97 & 54.48$\pm$1.00 & 53.49$\pm$0.99 & 51.20$\pm$0.97 \\
& & FedProx \cite{fedprox}& 58.41$\pm$0.94 & 57.15$\pm$0.98 & 56.51$\pm$0.96 & 55.61$\pm$0.98 & 54.51$\pm$1.00 & 53.29$\pm$0.97 & 51.78$\pm$1.00 \\
& & RoFL \cite{rofl}& 54.62$\pm$1.01 & 53.99$\pm$0.98 & 53.60$\pm$0.96 & 52.65$\pm$0.96 & 50.80$\pm$0.99 & 50.61$\pm$0.96 & 46.69$\pm$0.98 \\
& & RHFL \cite{RHFL}& 29.16$\pm$1.69 & 27.90$\pm$2.24 & 26.05$\pm$3.11 & 23.74$\pm$4.22 & 22.00$\pm$5.18 & 19.18$\pm$7.05 & 16.65$\pm$7.85 \\
& & FedLSR \cite{fedlsr}& 49.61$\pm$1.03 & 47.98$\pm$1.00 & 47.52$\pm$1.01 & 46.68$\pm$0.96 & 45.07$\pm$1.01 & 43.73$\pm$1.03 & 41.63$\pm$0.99 \\
& & FedCorr \cite{fedcorr}& 53.48$\pm$0.94 & 52.68$\pm$0.99 & 52.30$\pm$1.04 & 51.56$\pm$0.95 & 51.59$\pm$0.98 & 51.22$\pm$0.97 & 50.56$\pm$1.04 \\
& & FedNed \cite{fedned}& 57.53$\pm$0.95 & 56.68$\pm$1.00 & 57.30$\pm$1.00 & 57.03$\pm$0.96 & 56.50$\pm$0.97 & 56.43$\pm$0.98 & 56.26$\pm$0.99 \\
& & FedELC\cite{FedELC}& 58.02$\pm$1.01 & 58.04$\pm$1.00 & 57.88$\pm$0.98 & 57.49$\pm$0.97 & 57.57$\pm$1.00 & 57.21$\pm$0.97 & 57.05$\pm$0.97 \\
& & FedNoRo\cite{fednoro}& \underline{59.76$\pm$1.04} & \underline{59.61$\pm$1.01} & \underline{58.80$\pm$1.00} & \underline{58.72$\pm$0.97} & \underline{57.88$\pm$0.93} & \underline{57.58$\pm$0.99} & \underline{57.19$\pm$0.98} \\
& & \textbf{FedSIR (Ours)} & \textbf{60.42$\pm$0.98} & \textbf{59.81$\pm$0.95} & \textbf{59.44$\pm$0.94} & \textbf{58.81$\pm$0.97} & \textbf{58.91$\pm$0.94} & \textbf{58.27$\pm$1.00} & \textbf{58.11$\pm$1.01} \\

\cmidrule(lr){2-10}
& 5 & Pruning & \multicolumn{7}{c}{56.87$\pm$1.02} \\
& 10 & FedAvg & \multicolumn{7}{c}{63.05$\pm$0.92} \\

\bottomrule
\end{tabular}
\end{adjustbox}
\end{table*}

\section{Relabeling Strategy}
\begin{figure}[H]
    \centering
    \includegraphics[width=\linewidth]{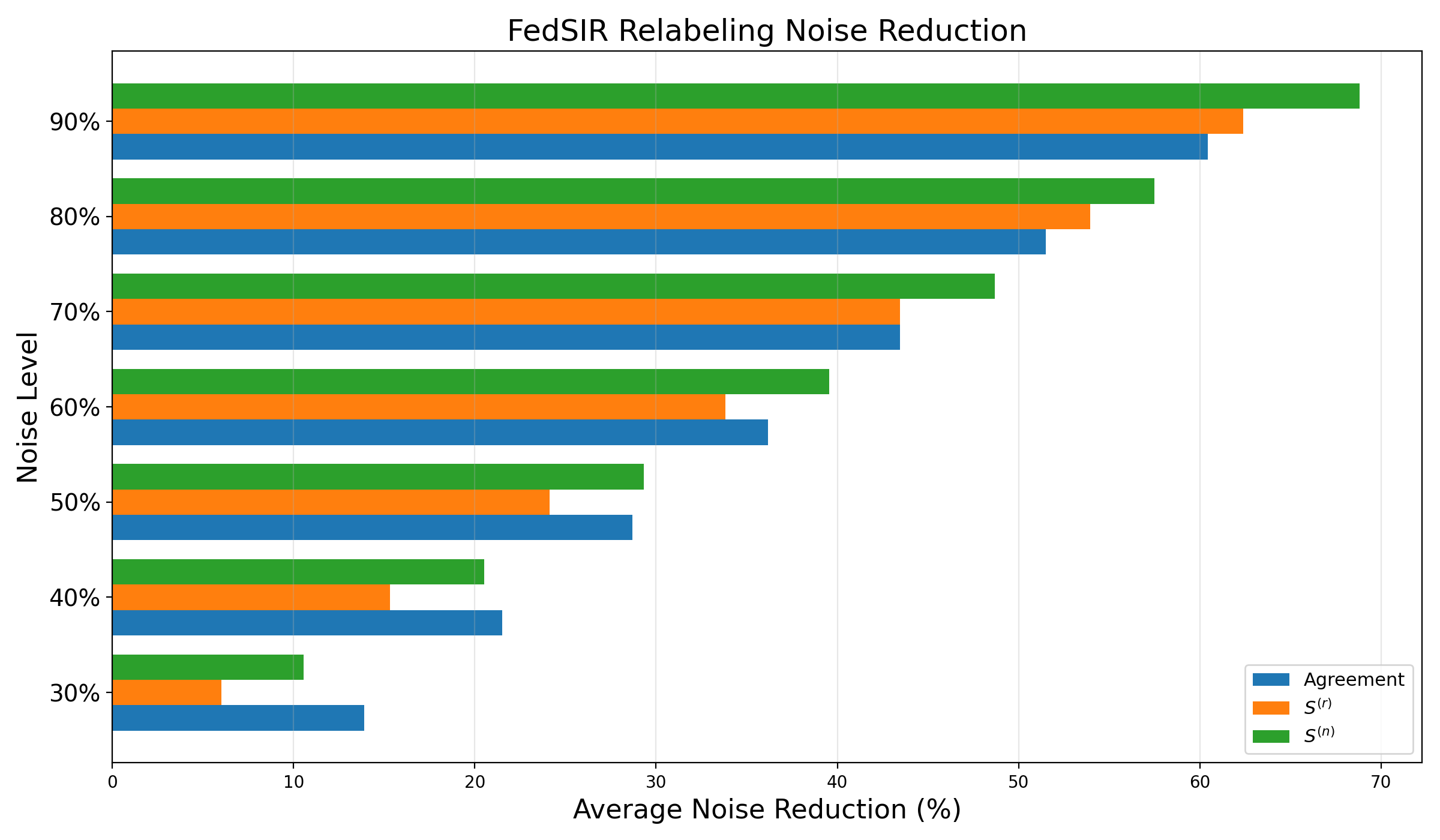}
    \caption{Average relabeling noise reduction under different symmetric noise levels for three relabeling variants: using only the dominant-direction score \(S^{(r)}\), using only the residual-subspace score \(S^{(n)}\), and the proposed agreement-based strategy.}
    \label{fig:relabel_noise_reduction}
\end{figure}

To analyze the role of the proposed relabeling rule, we compare three variants of the spectral correction mechanism used in Stage~II:
\begin{itemize}
\item $S^{(r)}$: labels are reassigned according to the dominant-direction alignment score: 
\begin{align*}
\hat{y}^{(r)}_i = \arg\max_c S^{(r)}(i,c).
\end{align*}

\item $S^{(n)}$: labels are determined using the residual-subspace projection score:
\begin{align*}
\hat{y}^{(n)}_i = \arg\min_c S^{(n)}(i,c).
\end{align*}
\item Agreement rule: a relabel is accepted only when the scores from $S^{(r)}$ and $S^{(n)}$ coincide:
$\hat{y}^{(r)}_i = \hat{y}^{(n)}_i$
\end{itemize}
Figure~\ref{fig:relabel_noise_reduction} reports the average reduction in label noise achieved by these strategies.

To evaluate the impact of these strategies on the final model performance, Table~\ref{tab:relabel_strategy} reports the test accuracy obtained using each relabeling variant.

\begin{table*}[t]
\centering
\caption{Effect of different relabeling strategies on CIFAR-10 under symmetric label noise. \textbf{Bold} indicates the best result and \underline{underline} indicates the second-best result.}
\label{tab:relabel_strategy}
\begin{adjustbox}{width=\textwidth}
\begin{tabular}{c c l c c c c c c c}
\toprule
\multicolumn{3}{c}{} & \multicolumn{7}{c}{Symmetric Noise} \\
\cmidrule(lr){4-10}
$\alpha$ & \# Clean Clients & Method
& 30\% & 40\% & 50\% & 60\% & 70\% & 80\% & 90\% \\
\midrule

\multirow{3}{*}{1} & \multirow{3}{*}{3}
& $S^{(r)}$ 
& \underline{84.73$\pm$0.70} 
& \underline{84.64$\pm$0.72} 
& \underline{84.45$\pm$0.71} 
& 84.07$\pm$0.73 
& \underline{84.31$\pm$0.68} 
& 83.95$\pm$0.70 
& \underline{84.06$\pm$0.70} \\

& & $S^{(n)}$ 
& 84.43$\pm$0.70 
& 84.21$\pm$0.70 
& 84.15$\pm$0.71 
& \underline{84.34$\pm$0.72} 
& 84.18$\pm$0.72 
& \underline{84.25$\pm$0.71} 
& 83.89$\pm$0.72 \\

& & Agreement
& \textbf{85.21$\pm$0.70} 
& \textbf{85.28$\pm$0.68} 
& \textbf{84.86$\pm$0.70} 
& \textbf{84.68$\pm$0.71} 
& \textbf{84.54$\pm$0.73} 
& \textbf{84.35$\pm$0.74} 
& \textbf{84.51$\pm$0.72} \\

\bottomrule
\end{tabular}
\end{adjustbox}
\end{table*}

\end{document}